\documentclass{article}
\usepackage{spconf,amsmath,graphicx}
\usepackage{subfig}

\title{Zoom in to the Details of Human-centric Videos}
\name{Guanghan Li$^{1}$, Yaping Zhao$^{1}$
 , Mengqi Ji$^{1}$,  Xiaoyun Yuan$^{1,2}$, Lu Fang$^{1}$\thanks{This work is supported in part by Natural Science Foundation of China (NSFC) under contract No. 61722209 and 6181001011, in part by Shenzhen Science and Technology Research and Development Funds (JCYJ20180507183706645, ZDYBH201900000002).}}
\address{$^1$ Tsinghua University\\
$^2$ The Hong Kong University of Science and Technology}
	
\begin{document}

\maketitle

\begin{abstract}

Presenting high-resolution (HR) human appearance is always critical for the human-centric videos. However, current imagery equipment can hardly capture HR details all the time. Existing super-resolution algorithms barely mitigate the problem by only considering universal and low-level priors of image patches. In contrast, our algorithm is under bias towards the human body super-resolution by taking advantage of high-level prior defined by HR human appearance.
Firstly, a motion analysis module extracts inherent motion pattern from the HR reference video to refine the pose estimation of the low-resolution (LR) sequence.
Furthermore, a human body reconstruction module maps the HR texture in the reference frames onto a 3D mesh model.
Consequently, the input LR videos get super-resolved HR human sequences are generated conditioned on the original LR videos as well as few HR reference frames.
Experiments on an existing dataset and real-world data captured by hybrid cameras show that our approach generates superior visual quality of human body compared with the traditional method. 

\end{abstract}


\vspace{-10pt}
\section{Introduction}
\label{sec:intro}
\vspace{-5pt}

Presenting high-resolution (HR) human appearance is always critical for the human-centric videos. Due to the restriction of the imagery equipment that is unable to capture HR human details all the time, video super-resolution (SR) algorithms are widely exploited to recover HR details from low-resolution (LR) frames. 

Obviously, super-resolution is inherently ill-posed owing to the many-to-one possible solutions mapping from HR images to LR ones. Therefore, extra prior knowledge is crucial to regularize the valid space. Even though the universal and low level image priors, such as stationary structure and sparse distribution, makes the traditional single-image SR (SISR) possible, the high dimension of valid space leads to blurry results.

Compared with SISR, recent Reference-based super-resolution (RefSR) methods \cite{refsr_zheng2017learning, refsr_zheng2018crossnet} dramatically compress the valid space by considering an extra HR reference frame as prior knowledge. Because it learns to warp the HR details from the reference frame to the LR template across large resolution gap up to 8x, apparently the ubiquitous large motion and occlusion in the reference frame worsens the warping estimation and performance. Therefore, it is not feasible to produce vivid details for the human-centric SR with large non-rigid deformation.
\begin{figure}
    \includegraphics[width = 0.48\textwidth]{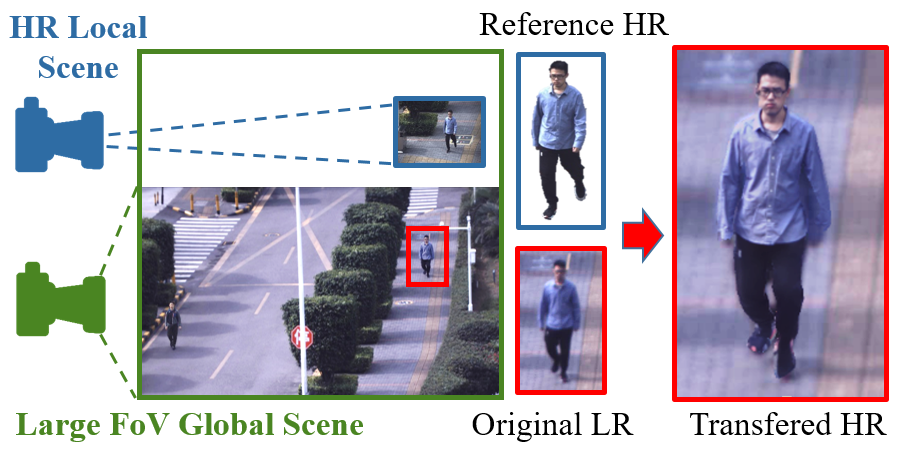}
    \caption{Illustration of the human-centric RefSR that transfers the high-definition human body details onto low resolution video.}
    \label{fig:teaser}
    \vspace{-10pt}
\end{figure}

\begin{figure*}[!t]
    \centering
    \includegraphics[width = 0.95\textwidth]{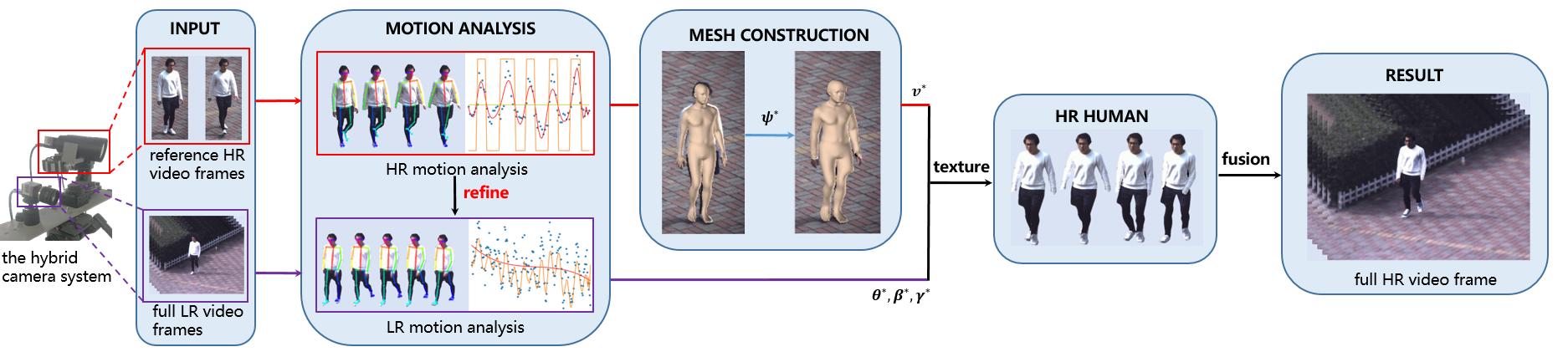}
    \caption{The \textit{HumanHR} pipeline to perform RefSR of human-centric video conditioned on an LR human video sequence as well as few HR frames of the corresponding human body. }
    \label{fig:pipeline}
\end{figure*}
In contrast, for the first time, this paper proposes a novel reference based human-centric video super-resolution algorithm, denoted as ``\emph{HumanSR}'' to infer HR  video conditioned on an LR human video sequence as well as few HR frames of the corresponding human body. 
Specifically, a motion analysis module extracts the intrinsic low-resolution (LR) pose refined by the HR motion pattern. The refinement takes advantage of the low rank characteristics of human motion, i.e., the periodic motion and invariant appearance.
Additionally, a human body reconstruction module fits a 3D human model and maps the HR texture onto a 3D mesh model.
Consequently, HR human sequences are generated conditioned on the original LR videos as well as few HR reference frames.
The experiments on the dataset MPII\cite{andriluka20142d} demonstrate the tremendous performance gap between HumanSR and the state-of-the-art methods in terms of visual quality. 
The experiment on real captured pedestrian video shows that the proposed approach generates much superior RefSR results compared with state-of-the-art solutions.


\vspace{-10pt}
\section{RELATED WORK}
\label{sec:related_work}
\vspace{-5pt}
\noindent{\textbf{Super-resolution.}}
In the early days, manually designed priors, such as sparsity prior \cite{yang2012coupled} and exemplar patches \cite{glasner2009super}, were utilized.
Recently, deep learning methods boost the SISR performance. Dong \emph{et al.} first proposed SRCNN \cite{nnsr_dong2014learning}, a simple $3$-layer ConvNet, to recover HR details. With the increasing model capacity of the deep neural networks, the SISR performance has been rapidly improved. Additionally, Tao \emph{et al.} extended the super-resolution to videos by fusing multiple frames to reveal image details\cite{videosr_tao2017detail}. Nevertheless, the performance of SISR is limited by the universal and low-level image prior leading to large valid-solution space and blurry results. 

Recent works \cite{refsr_boominathan2014improving}\cite{refsr_zheng2017learning}\cite{refsr_zheng2018crossnet} considers additional HR images from different viewpoints or timestamp to assist super-resolving the LR input, which forms a new kind of SR method called RefSR. The imported HR reference frame, middle-level prior, lets RefSR achieves promising performance. CrossNet \cite{refsr_zheng2018crossnet} dramatically improves RefSR by learning the cross-scale warping from an HR frame to an LR template. However, it cannot deal with the region with large deformation, e.g., human body. In contrast, for human-centric video-SR, our method takes advantage of pose and motion regularization of human body as high-level prior.

\vspace{5pt}
\noindent{\textbf{Human Body Reconstruction.}}
3D Human body reconstruction from 2d RGB images can be classified into two categories: by universal multi-view sterepsis \cite{cheung2003visual} and by human body specific prior knowledge \cite{3d_from_video_alldieck2018video}\cite{smpl_loper2015smpl}\cite{keepitsmpl_bogo2016keep}\cite{hmr_kanazawa2018end}. Loper \emph{et al.} proposed SMPL\cite{smpl_loper2015smpl}, a general 3d human template for human reconstruction, deformation, etc. Bogo \emph{et al.} proposed to build the SMPL model from 2d human image and skeleton\cite{keepitsmpl_bogo2016keep}. However, the deformation of the reconstructed 3D human model is guided by high-resolution frame without scale gap.

\vspace{-10pt}
\section{Proposed Method}
\label{sec:method}
\vspace{-5pt}
In this section, we present the pipeline of cross-scale human-centric detailed recovery illustrated in Fig.\ref{fig:pipeline}. 
Firstly, the SMPL parameters are estimated and refined by analyzing both LR and HR video sequences.
Furthermore, a non-rigid 3D human model is constructed to cover the human silhouette.
Consequently, the dynamic human model with the 2D HR human details are rendered onto the original LR video. 

\vspace{-5pt}
\subsection{Motion Analysis}
\label{sec:analysis}

\noindent{\textbf{Pose Estimation.}}
In this paper, we adopt SMPL\cite{smpl_loper2015smpl} model to represent 3D body. The SMPL model is define as a function $M(\beta,\theta,\gamma)$, parameterized by shape $\beta$, pose $\theta$, and translation $\gamma$. The pose $\theta = [\omega_1, ..., \omega_K]^T$ is defined by a skeleton rig with $K = 23$ joints. Hence a pose $\theta$ has ${3 \times 23 + 3 = 72}$ parameters, $3$ for each joint and $3$ for the root orientation.

First, we extended the widely used SMPL parameters estimation method 
\cite{keepitsmpl_bogo2016keep} by introducing mask and temporal information. The SMPL parameters for HR $\left\{\theta_{HR}, \beta_{HR}, \gamma_{HR}\right\}$ and LR $\left\{\theta_{LR}, \beta_{LR}, \gamma_{LR}\right\}$ is both estimated by the following objective function:
\begin{equation}
    \theta^*, \beta^*, \gamma^*  = \mathop{\arg\min}_{\theta, \beta, \gamma}\ \ \omega_{2d}E_{2d} + \omega_{3d}E_{3d} + \omega_{m}E_m + \omega_{S}E_S.
\end{equation}

The 2d joint term $E_{2d}$ is derived from the period method\cite{keepitsmpl_bogo2016keep} which the 2d joint detected by OpenPose\cite{2d_pose_cao2017realtime}. The mask term $E_m$ enforces a dense correspondence between 3d model and image, which is defined as:  
\begin{equation}
    E_m =  M_{3d} \odot \overline{M_{2d}} + \lambda \overline{M_{3d}} \odot M_{2d},
\end{equation}
where $M_{3d}$ is the projected masks from SMPL model, $M_{2d}$ is the human mask estimated from the RGB image using mask-rcnn\cite{mask_rcnn_he2017mask}. $\overline{M_{3d}}$ and $\overline{M_{2d}}$ represent their inverse masks. $\lambda$ is the weight, and $\odot$ represents the element-wise product of two masks.

The 3d term $E_{3d}$ is aiming to solve the depth ambiguity:
\begin{equation}
    E_{3d} = \sum_{i=1}^{\theta}\|\theta - \theta_{hmr}\|^2_2,
\end{equation}
where $\theta$ is the pose parameter in SMPL, and $\theta_{hmr}$ is the parameter given by HMR\cite{hmr_kanazawa2018end}.

Finally, we add a temporal smooth term and a mesh smooth term:  
\begin{equation}
    E_S = \lambda_1 \sum_{i=1}^{N}\left\| J^{3d}_{i, t} - J^{3d}_{i, t+1} \right\|_2^2  + \lambda_2 \sum_{i=1}^{m}\left\| v_i^f - v_i\right\|_2^2,
\end{equation}
where $J^{3d}_{i, t}$ and $J^{3d}_{i, t+1}$ is the 3d model joint position in t-frame and (t+1)-frame respectively. The latter is derived from the optical flow constraint\cite{xiang2018monocular}. Note that the $\lambda_2$=0 in LR video due to the low imaging quality.

The aforementioned method may still produce jitters due to the blur and low quality. Furthermore, the hyperparameters are difficult to determine. Therefore, we estimate the $\theta_{R}$ from $\theta_{LR}$ and the reference $\theta_{HR}$ as the final $\theta_{LR}^*$. The method to calculate the $\theta_{R}$ is described below. 

\vspace{5pt}
\noindent{\textbf{Motion Refinement.}}
We propose an algorithm that refines the $\theta_{LR}$ based on time series analysis\cite{brown2004smoothing}, especially seasonality analysis\cite{hylleberg1992modelling}.
The key insight behind this is that most human actions are approximately repetitive, such as walking, running, physical exercise, etc. We choose the $\theta_{1,1}$ as an example, which represents the 1st axis-angle of the 1st joint in the SMPL model (the ankle joint of the human).



\begin{figure}[]
    \subfloat[]{\includegraphics[width = 0.24\textwidth]{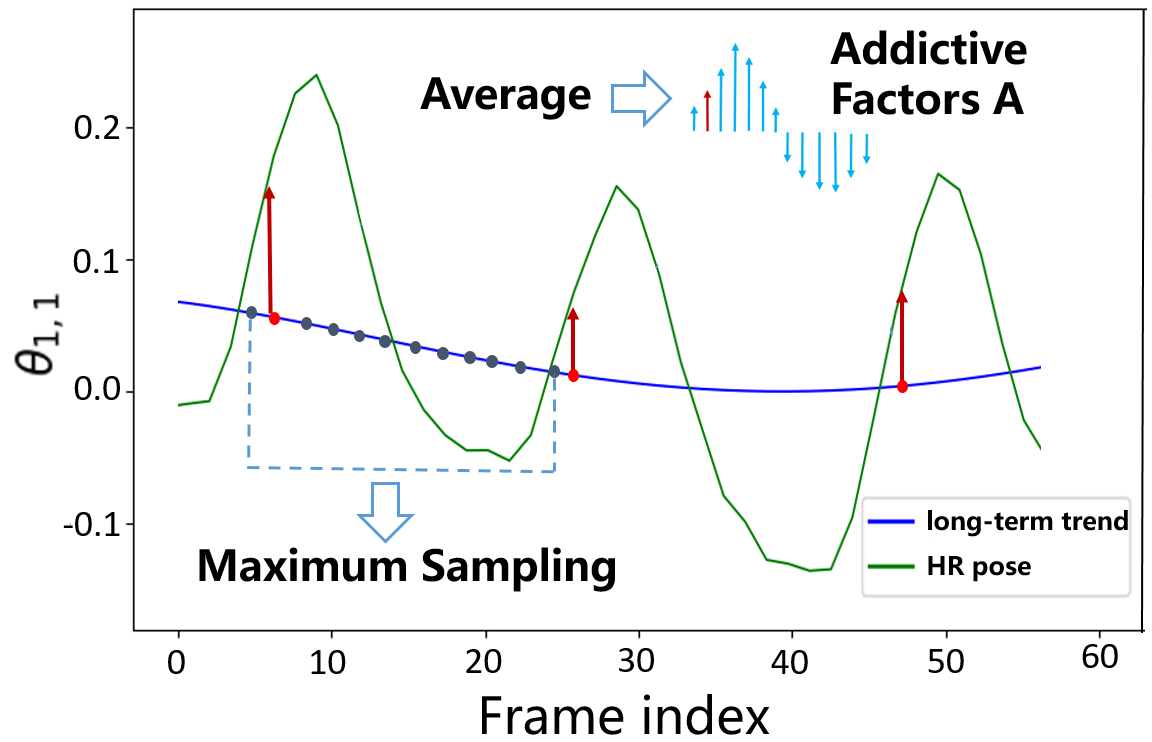}}
    \subfloat[]{\includegraphics[width = 0.24\textwidth]{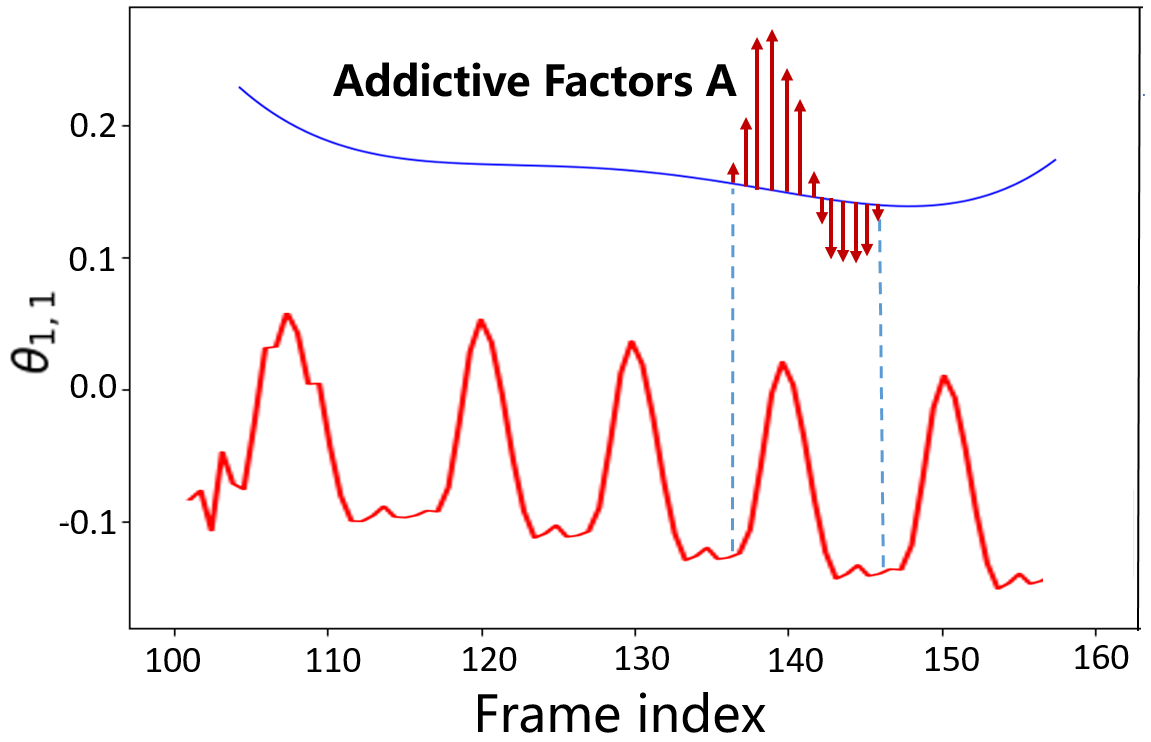}}
    \caption{(a) The addictive factors are calculated through comparing the LR long-term trend and $P_{HR}$. (b) The final motion is refined by add the addictive factors to LR long-term trend.}
    \label{fig:motion-refine}
\end{figure}


Firstly, auto-correlation function (ACF) is utilized to identify the seasonality\cite{plosser1979short}. Note that though the human motion is regularly repetitive, the period is not fixed but around a constant value. With the seasonality, we build an additive time-series model\cite{maravall1987prototypical} by extracting a long-term trend $\mathbf{L}$ from LR poses sequence with polynomial fitting based on the least-square method. 

In order to smooth the HR and LR poses sequences, we use the moving average to generate $P_{HR}$ and $P_{LR}$. After that, the different periods $T_{LR}$ can be extracted by finding the crossover points of $\mathbf{L}$ and $P_{LR}$. We then sample each period of LR sequence with the difference between $P_{HR}$ and $\mathbf{L}$ uniformly, and the number of sampling points in each period is the maximum period $T_{LR}^M$ in $T_{LR}$. Notes that the non-integer samplings are interpolated by linear method. For stability, all sampling values are averaged by the period to generate the addictive factors $\mathbf{A}$ whose length is $T_{LR}^M$. Finally, we add the $\mathbf{A}$ back to each period of $\mathbf{L}$ by linear interpolation to generate the final LR sequence. The process preserves the LR trend as well as the details in the shorter HR sequences, as shown in Fig.\ref{fig:motion-refine}(a)(b).




\vspace{-10pt}
\subsection{mesh construction}
\label{sec:construction}
\vspace{-5pt}
We now have the HR SMPL parameters, but the model may not adapt to the individual human shape. So we reconstruct a non-rigid 3d mesh to fit the human contour in 2d image. We firstly select a frame with minimal self-occlusion to obtain the human texture. The key frame is found by projecting the SMPL to 2d image plane, and check the overlapping regions of the body parts. To seek some corresponding point pairs between projected SMPL contours $p^{S}$ and mask contours $p^{m}$ in the key frame, we define the following objective function:
\begin{equation}
    \psi^* = \mathop{\arg\min}_{\psi} \ \ \sum_{i=1}^{N} \left\| p^{m}_i - p^{S}_{\psi[i]}\right\|^2_2 +  \lambda \sum_{i=1}^{N-1} \| \psi[i+1] - \psi[i]\|_2^2,
    \label{eq:psi}
\end{equation}
where the former penalizes the discrepancy between each mask contour $p^{m}_i$ and corresponding projected SMPL contour $p^{S}_{\psi[i]}$, the $\psi[i]$ maps the $i_{th}$ in human mask contour to the index of SMPL contours. The latter is the smooth term with weight $\lambda$ which avoids the jump between the $\psi[i]$ and $\psi[i+1]$. The $\psi^*$ can be solved efficiently using $\alpha$-expansion\cite{gco_delong2012fast} and some misalignment can be reduced after optimization. Note that the common segmentation method, just like Mask-rcnn\cite{mask_rcnn_he2017mask}, may produce some artifacts on edge. So we use the segmentation network\cite{zheng2015conditional} based on conditional random fields, and then the Dense CRF\cite{krahenbuhl2011efficient} is used to refine the contours.


With the corresponding contours pairs, the deformed human model can be estimated by the following objective function:
\begin{equation}
    v^* = \mathop{\arg\min}_{v} \ \ \sum_{i=1}^{N} \left\| p^{m}_i - p^{S}_{\psi[i]}\right\|^2_2 + \sum_{i=1}^N \omega_i \left\|L(v_i)-L_0(v_i)\right\|^2,
\end{equation}
the former has been mentioned in (\ref{eq:psi}) which enforces the point $p^{m}_i$ closed to $p_{\psi[i]}^{S}$. The latter is derived from a laplacian method\cite{3d_from_video_alldieck2018video} to keep the model surface smooth, the $L(v_i)$ convert the $i_{th}$ vertex $v_i$ on SMPL into laplacian coordination, $L_0(v_i)$ represents the initial laplacian coordination with non-deformed model. The full vertices $v^*$ can be estimated using L-BFGS-B optimizer. The texture of the deformed human model can be obtained by back projecting the image to 3d model. We reset the deformed model to original parameters to get the deformed SMPL template.  

Finally, the $\theta_{LR}^*, \beta_{LR}^*, \gamma_{LR}^*$ mentioned in \ref{sec:analysis} are applied to the deformed template. The model is rendered to generate the final human details.

\vspace{-10pt}
\section{Experiment}
\label{sec:experiment}

\vspace{-5pt}
\noindent{\textbf{Data Preparation.}}
The experiments are performed in both synthetic data and real-world data. Firstly, we build sets of synthetic data from MPII\cite{andriluka20142d}, which consists of massive human-centric images sequences with various human poses. Most sequences of MPII have about 40 frames, we set a single frame as the HR frame, the rest which are downsampled $\times8$ is regarded as LR frames. In order to evaluate our method in more general cases, we capture real outdoor scenes by our hybrid camera hardware setup, which contains a large-FoV global camera and a rotatable small-FoV local camera. In this system, we detect the pedestrians in the global camera in real time, and extract the high-resolution frames with the local camera. The dataset will get released to the public.

\begin{figure}
    \includegraphics[width = 0.48\textwidth]{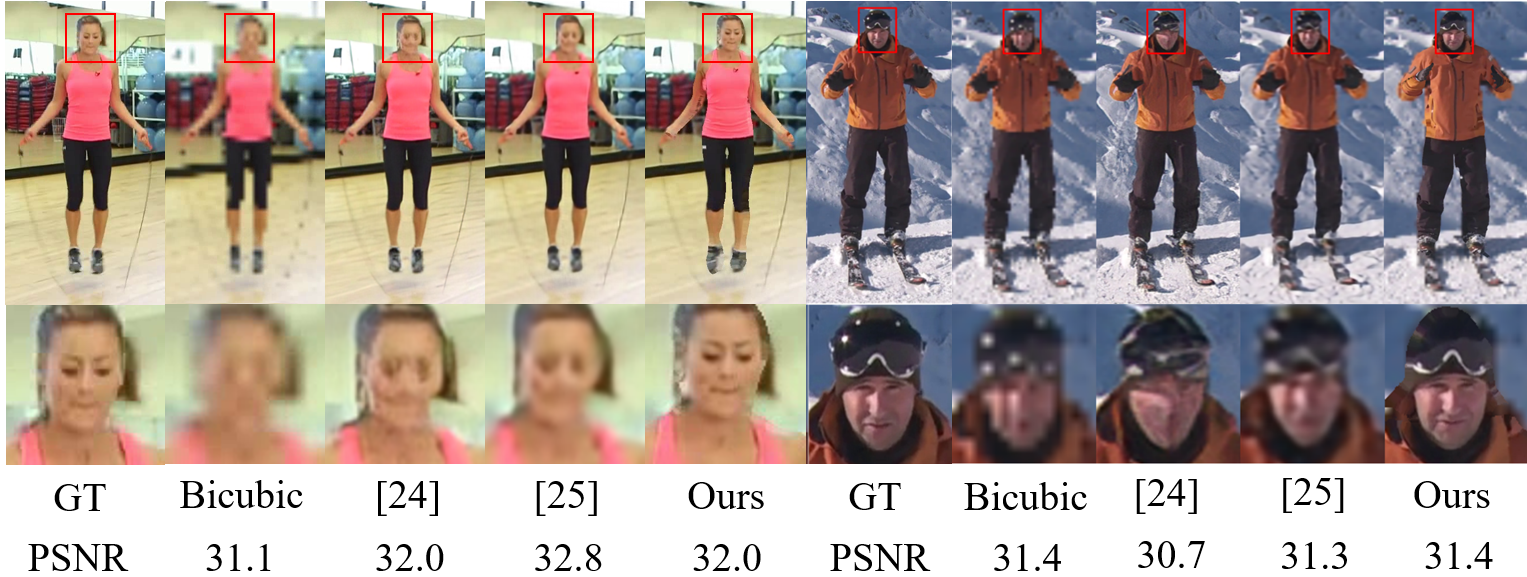}
    \caption{Results on $\times$ 8 RefSR in the synthesised MPII dataset. Our approach generates superior visual quality of human body despite of slightly lower PSNR, compared with the methods: Bicubic, Wang \emph{et al.}\cite{ESRGAN_wang2018esrgan}, Tao \emph{et al.}\cite{tao2017detail}.}
    \label{fig:mpii_ex}
\end{figure}

\begin{figure}
    \includegraphics[width = 0.45\textwidth]{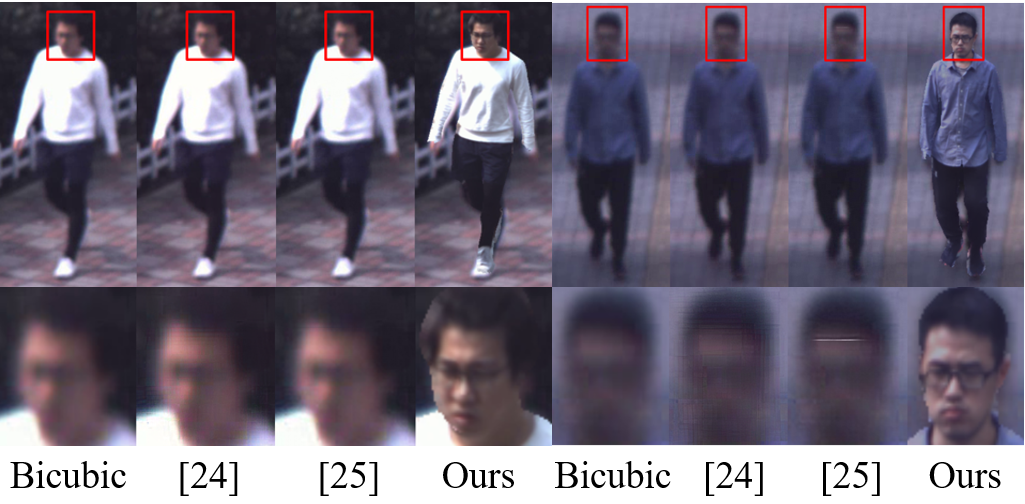}
    \caption{Real-world data captured by our hybrid-camera hardware setup. Our results is much appealling compared with the traditional the methods: Bicubic, Wang \emph{et al.}\cite{ESRGAN_wang2018esrgan}, Tao \emph{et al.}\cite{tao2017detail}. They are $\times$ 4 super-resolved.}
    \label{fig:real_ex}
\end{figure}

\vspace{5pt}
\noindent{\textbf{Comparison with Details Recovery.}}
We compare our method with Wang \emph{et al.}\cite{ESRGAN_wang2018esrgan} and Tao \emph{et al.}\cite{tao2017detail} respectively. In the case of synthetic data, shown in Fig.\ref{fig:mpii_ex}, \cite{ESRGAN_wang2018esrgan}\cite{tao2017detail} lead to blurry results. In contrast, our method produces appealing results despite of low PSNR, especially around the facial region. In the real-world dataset, the inputs of our method are ~30$\sim$~40 HR reference sequence and ~200$\sim$~600 LR sequence captured by the hybrid camera, and a super-resolved sequence is produced attached with the HR appearance. Because there is no ground truth frame in the real-world sequence, we qualitatively compare the results. As illustrated in Fig.\ref{fig:real_ex}, our method has more obvious superiority in real data. One advantage is that the performance of SR is bounded by the effectiveness of the cross-scale warping method rather than the scale difference.

\begin{figure}[!b]
    \includegraphics[width = 0.48\textwidth]{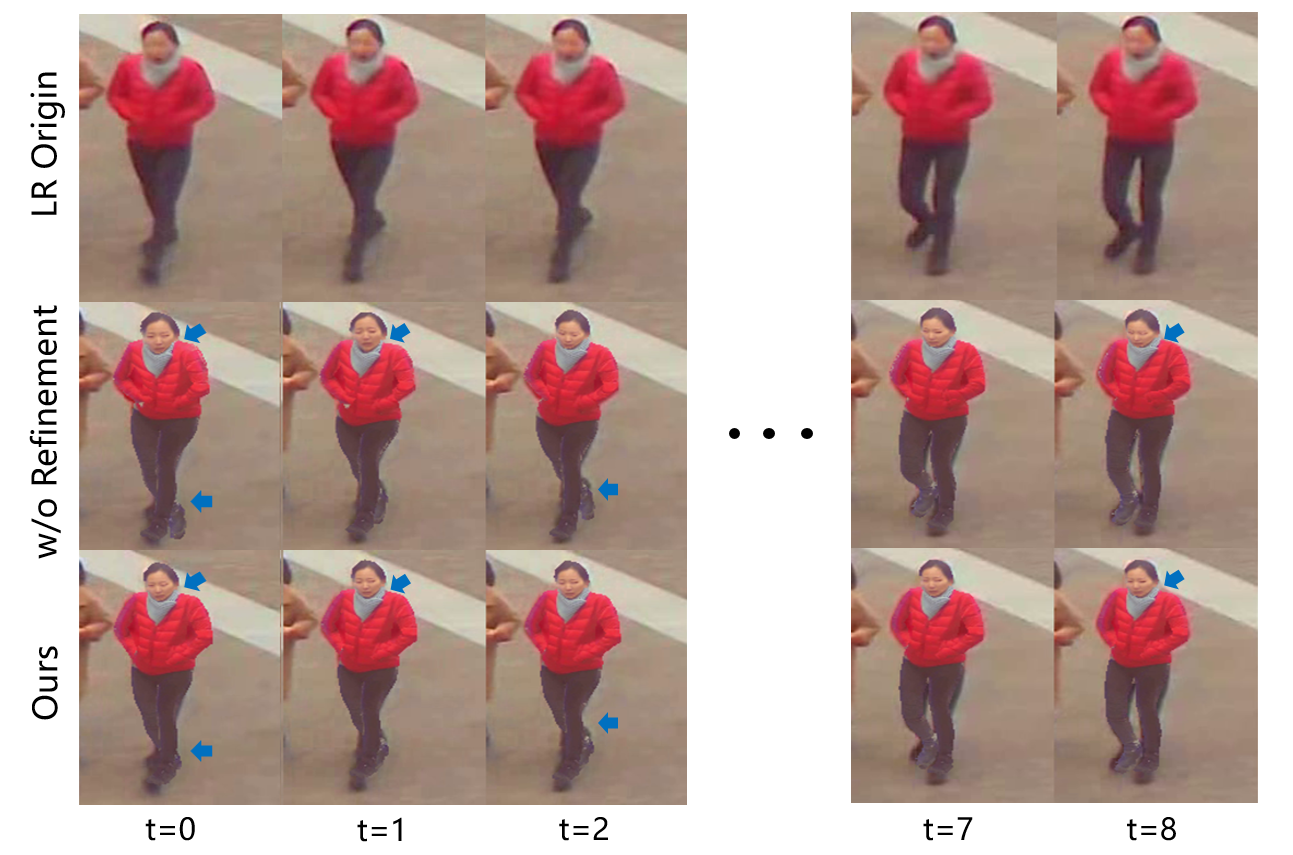}
    \caption{the proposed refinement approach observably eliminates the jitter artifact around the highly occluded region (blue arrow).}
    \label{fig:motion}
\end{figure}

\vspace{5pt}
\noindent{\textbf{Ablation Study.}}
To validate the effectiveness of the motion analysis module, we qualitatively evaluate it in a real-world sequence. As shown in Fig.\ref{fig:motion}, when the motion refinement is not enabled, temporal jitter significantly disrupts the visual quality of the human-centric video, especially on the highly occluded region (blue arrow). In contrast, the proposed refinement observably eliminates the jitter artifact.


\vspace{5pt}
\noindent{\textbf{Implementation Detail.}}
The full process of generating 3D motion is time-expensive with frame-by-frame,
so we adopt the batch optimization with multi-frames to speed up the optimizing time. As the length of 600 frames, it needs 4-5 seconds per-frame with frame-by-frame optimizing, compared with about 1 second per-frame with the batch optimizing. All of them use L-BFGS-B as optimizing algorithm, running in Nvidia GTX1070 with TensorFlow. Because of the large batch size, it comes along with the declining accuracy due to the difficult convergence with a large number of parameters. Benefited from our motion refinement, the lost accuracy can be offset well.

\vspace{-10pt}
\section{Conclusion}
\label{ssec:conclusion}
\vspace{-5pt}
For the first time, we present a novel reference based super-resolution algorithm for human-centric video. The imported prior knowledge can provide high-level regularization. Specifically, a human body reconstruction module maps the HR texture onto a 3D mesh model. Then, a motion analysis module extracts the intrinsic pedestrian motion pattern for natural human motion refinement before the HR human sequences are generated.
While the results of our method show slightly lower PSNR compared to traditional methods, the super-resolved human-centric video is more appealing when viewed by human.
\bibliographystyle{IEEEbib}
\small{\bibliography{refs}}

\begin{thebibliography}{10}

\bibitem{refsr_zheng2017learning}
Haitian Zheng, Mengqi Ji, Haoqian Wang, Yebin Liu, and Lu~Fang,
\newblock ``Learning cross-scale correspondence and patch-based synthesis for
  reference-based super-resolution,''
\newblock BMVC, 2017.

\bibitem{refsr_zheng2018crossnet}
Haitian Zheng, Mengqi Ji, Haoqian Wang, Yebin Liu, and Lu~Fang,
\newblock ``Crossnet: An end-to-end reference-based super resolution network
  using cross-scale warping,''
\newblock in {\em Proceedings of the European Conference on Computer Vision
  (ECCV)}, 2018, pp. 88--104.

\bibitem{andriluka20142d}
Mykhaylo Andriluka, Leonid Pishchulin, Peter Gehler, and Bernt Schiele,
\newblock ``2d human pose estimation: New benchmark and state of the art
  analysis,''
\newblock in {\em Proceedings of the IEEE Conference on computer Vision and
  Pattern Recognition}, 2014, pp. 3686--3693.

\bibitem{yang2012coupled}
Jianchao Yang, Zhaowen Wang, Zhe Lin, Scott Cohen, and Thomas Huang,
\newblock ``Coupled dictionary training for image super-resolution,''
\newblock {\em IEEE transactions on image processing}, vol. 21, no. 8, pp.
  3467--3478, 2012.

\bibitem{glasner2009super}
Daniel Glasner, Shai Bagon, and Michal Irani,
\newblock ``Super-resolution from a single image,''
\newblock in {\em 2009 IEEE 12th international conference on computer vision}.
  IEEE, 2009, pp. 349--356.

\bibitem{nnsr_dong2014learning}
Chao Dong, Chen~Change Loy, Kaiming He, and Xiaoou Tang,
\newblock ``Learning a deep convolutional network for image super-resolution,''
\newblock in {\em European conference on computer vision}. Springer, 2014, pp.
  184--199.

\bibitem{videosr_tao2017detail}
Xin Tao, Hongyun Gao, Renjie Liao, Jue Wang, and Jiaya Jia,
\newblock ``Detail-revealing deep video super-resolution,''
\newblock in {\em Proceedings of the IEEE International Conference on Computer
  Vision}, 2017, pp. 4472--4480.

\bibitem{refsr_boominathan2014improving}
Vivek Boominathan, Kaushik Mitra, and Ashok Veeraraghavan,
\newblock ``Improving resolution and depth-of-field of light field cameras
  using a hybrid imaging system,''
\newblock in {\em 2014 IEEE International Conference on Computational
  Photography (ICCP)}. IEEE, 2014, pp. 1--10.

\bibitem{cheung2003visual}
German~KM Cheung, Simon Baker, and Takeo Kanade,
\newblock ``Visual hull alignment and refinement across time: A 3d
  reconstruction algorithm combining shape-from-silhouette with stereo,''
\newblock in {\em 2003 IEEE Computer Society Conference on Computer Vision and
  Pattern Recognition, 2003. Proceedings.} IEEE, 2003, vol.~2, pp. II--375.

\bibitem{3d_from_video_alldieck2018video}
Thiemo Alldieck, Marcus Magnor, Weipeng Xu, Christian Theobalt, and Gerard
  Pons-Moll,
\newblock ``Video based reconstruction of 3d people models,''
\newblock in {\em Proceedings of the IEEE Conference on Computer Vision and
  Pattern Recognition}, 2018, pp. 8387--8397.

\bibitem{smpl_loper2015smpl}
Matthew Loper, Naureen Mahmood, Javier Romero, Gerard Pons-Moll, and Michael~J
  Black,
\newblock ``Smpl: A skinned multi-person linear model,''
\newblock {\em ACM transactions on graphics (TOG)}, vol. 34, no. 6, pp. 248,
  2015.

\bibitem{keepitsmpl_bogo2016keep}
Federica Bogo, Angjoo Kanazawa, Christoph Lassner, Peter Gehler, Javier Romero,
  and Michael~J Black,
\newblock ``Keep it smpl: Automatic estimation of 3d human pose and shape from
  a single image,''
\newblock in {\em European Conference on Computer Vision}. Springer, 2016, pp.
  561--578.

\bibitem{hmr_kanazawa2018end}
Angjoo Kanazawa, Michael~J Black, David~W Jacobs, and Jitendra Malik,
\newblock ``End-to-end recovery of human shape and pose,''
\newblock in {\em Proceedings of the IEEE Conference on Computer Vision and
  Pattern Recognition}, 2018, pp. 7122--7131.

\bibitem{2d_pose_cao2017realtime}
Zhe Cao, Tomas Simon, Shih-En Wei, and Yaser Sheikh,
\newblock ``Realtime multi-person 2d pose estimation using part affinity
  fields,''
\newblock in {\em Proceedings of the IEEE Conference on Computer Vision and
  Pattern Recognition}, 2017.

\bibitem{mask_rcnn_he2017mask}
Kaiming He, Georgia Gkioxari, Piotr Doll{\'a}r, and Ross Girshick,
\newblock ``Mask r-cnn,''
\newblock in {\em Proceedings of the IEEE international conference on computer
  vision}, 2017, pp. 2961--2969.

\bibitem{xiang2018monocular}
Donglai Xiang, Hanbyul Joo, and Yaser Sheikh,
\newblock ``Monocular total capture: Posing face, body, and hands in the
  wild,''
\newblock {\em arXiv preprint arXiv:1812.01598}, 2018.

\bibitem{brown2004smoothing}
Robert~Goodell Brown,
\newblock {\em Smoothing, forecasting and prediction of discrete time series},
\newblock Courier Corporation, 2004.

\bibitem{hylleberg1992modelling}
Svend Hylleberg,
\newblock {\em Modelling seasonality},
\newblock Oxford University Press, 1992.

\bibitem{plosser1979short}
Charles~I Plosser,
\newblock ``Short-term forecasting and seasonal adjustment,''
\newblock {\em Journal of the American Statistical Association}, vol. 74, no.
  365, pp. 15--24, 1979.

\bibitem{maravall1987prototypical}
Agust{\'\i}n Maravall and David~A Pierce,
\newblock ``A prototypical seasonal adjustment model,''
\newblock {\em Journal of Time Series Analysis}, vol. 8, no. 2, pp. 177--193,
  1987.

\bibitem{gco_delong2012fast}
Andrew Delong, Anton Osokin, Hossam~N Isack, and Yuri Boykov,
\newblock ``Fast approximate energy minimization with label costs,''
\newblock {\em International journal of computer vision}, vol. 96, no. 1, pp.
  1--27, 2012.

\bibitem{zheng2015conditional}
Shuai Zheng, Sadeep Jayasumana, Bernardino Romera-Paredes, Vibhav Vineet,
  Zhizhong Su, Dalong Du, Chang Huang, and Philip~HS Torr,
\newblock ``Conditional random fields as recurrent neural networks,''
\newblock in {\em Proceedings of the IEEE international conference on computer
  vision}, 2015, pp. 1529--1537.

\bibitem{krahenbuhl2011efficient}
Philipp Kr{\"a}henb{\"u}hl and Vladlen Koltun,
\newblock ``Efficient inference in fully connected crfs with gaussian edge
  potentials,''
\newblock in {\em Advances in neural information processing systems}, 2011, pp.
  109--117.

\bibitem{ESRGAN_wang2018esrgan}
Xintao Wang, Ke~Yu, Shixiang Wu, Jinjin Gu, Yihao Liu, Chao Dong, Yu~Qiao, and
  Chen Change~Loy,
\newblock ``Esrgan: Enhanced super-resolution generative adversarial
  networks,''
\newblock in {\em Proceedings of the European Conference on Computer Vision
  (ECCV)}, 2018, pp. 0--0.

\bibitem{tao2017detail}
Xin Tao, Hongyun Gao, Renjie Liao, Jue Wang, and Jiaya Jia,
\newblock ``Detail-revealing deep video super-resolution,''
\newblock in {\em Proceedings of the IEEE International Conference on Computer
  Vision}, 2017, pp. 4472--4480.

\end{thebibliography}

\end{document}